\documentclass{article}
\usepackage{iclr2027_conference,times}

\iclrfinalcopy

\usepackage[utf8]{inputenc}
\usepackage{textcomp}

\usepackage{amsmath,amsfonts,bm}

\def\eqref#1{equation~\ref{#1}}

\def\1{\bm{1}}

\DeclareMathAlphabet{\mathsfit}{\encodingdefault}{\sfdefault}{m}{sl}
\SetMathAlphabet{\mathsfit}{bold}{\encodingdefault}{\sfdefault}{bx}{n}

\usepackage{hyperref}
\usepackage{url}
\usepackage{wrapfig}

\usepackage{xcolor}
\usepackage{fontawesome5}

\usepackage[most]{tcolorbox}
\tcbuselibrary{breakable,skins}
\usepackage{tabularx}
\usepackage{caption}
\usepackage{enumitem}

\usepackage{amsmath}
\usepackage{multirow}
\usepackage{amssymb}
\usepackage{amsthm}
\usepackage{booktabs}

\usepackage{adjustbox}

\definecolor{awFramePurple}{HTML}{4D2F82}
\definecolor{awParentBlue}{HTML}{24608C}
\definecolor{awEvolvedPurple}{HTML}{7D3194}

\definecolor{RtenPink}{HTML}{FCE8EB}
\definecolor{RtenGray}{HTML}{F2F2F2}

\usepackage{graphicx}
\usepackage{xcolor}
\usepackage{colortbl}
\usepackage{xspace}

\usepackage{hyperref}
\usepackage{fontawesome5}

\usepackage{array}

\newcolumntype{N}{%
  >{\fontencoding{OT1}\fontfamily{cmr}\selectfont}c%
}

\definecolor{PairShade}{HTML}{F1F1F1}
\definecolor{OursShade}{HTML}{FCE7E9}

\definecolor{lightblue}{RGB}{212, 235, 255}
\definecolor{orange}{RGB}{255, 105, 0}
\definecolor{lightgreen}{RGB}{177, 231, 171}
\definecolor{lightyellow}{RGB}{255, 255, 148}
\definecolor{darkgreen}{RGB}{50, 200, 50}
\definecolor{darkyellow}{RGB}{230, 200, 0}

\newcolumntype{P}[1]{>{\centering\arraybackslash}p{#1}}
\newcolumntype{L}[1]{>{\raggedright\let\newline\\\arraybackslash\hspace{0pt}}m{#1}}
\newcolumntype{R}[1]{>{\raggedleft\arraybackslash}p{#1}}

\title{HERA: Harness–Environment Co-Evolution for Reliable Agentic Abstention}

\author{
\textbf{Han Luo}$^{1,2}$\thanks{Equal contribution.} \quad
\textbf{Bingbing Wen}$^{1,*}$ \quad
\textbf{Guang Yang}$^{1}$ \\
~\textbf{Zora Zhiruo Wang}$^{3}$ \quad
\textbf{Pan Lu}$^{4}$ \quad
\textbf{Lucy Lu Wang}$^{1,5}$ \\[0.6em]
{\normalfont\small
$^{1}$University of Washington \quad
$^{2}$University of Leeds \quad
$^{3}$Carnegie Mellon University} \\
{\normalfont\small
$^{4}$Stanford University \quad
$^{5}$Allen Institute for AI} \\[0.4em]
{\normalfont\small
\hypersetup{hidelinks}
\href{https://hera-bench.github.io/}{%
  \textcolor[HTML]{4B2E83}{%
    \faGlobeAmericas\enspace
    \nolinkurl{hera-bench.github.io}}}
}
}

\newcommand{\framework}{HERA\xspace}
\newcommand{\dataset}{\textsc{hera-bench}\xspace}

\usepackage{titlesec}

\titlespacing*{\section}{0pt}{1.5ex plus 0.5ex minus 0.2ex}{0.7ex}
\titlespacing*{\subsection}{0pt}{1.2ex plus 0.4ex minus 0.2ex}{0.5ex}
\begin{document}
\maketitle
\fancyhead{}
\renewcommand{\headrulewidth}{0pt}

\begin{abstract}

Large language model (LLM) agents are increasingly capable of acting in complex tool-use environments, yet they often fail to recognize when tasks are infeasible and no valid solution exists. Recent work has formalized this reliability gap as the problem of \emph{agentic abstention}, and existing approaches typically optimize a model or agent harness against a fixed set of tasks, leading to limited generalization to unseen failure modes. We introduce \framework, a framework for harness--environment co-evolution for agentic abstention. \framework consists of (i) a pipeline to automatically construct verifiable pairs of feasible and infeasible tasks by applying controlled environment mutations that transform solvable tasks into cases requiring abstention, and (ii) a co-evolution procedure in which performance failures on previous tasks are used to drive harness adaptation and generate new execution environments and tasks geared towards previous weaknesses. 
On held-out evaluation tasks, an evolved harness from \framework improves abstention accuracy from 61.7\% to 83.3\% while improving feasible-task completion from 68.3\% to 76.7\%, achieving the highest abstention and feasible-task completion among the compared methods. 
The resulting best harness transfers across 19 other LLMs, improving abstention accuracy by 15.3 percentage points on average 
without any model-specific optimization, and enabling smaller models to match the performance of more powerful models at an estimated 85\% lower cost.

\end{abstract}

\section{Introduction}

Large language model (LLM) agents are increasingly capable of solving complex tasks through multi-step interactions with tools and external environments \citep{merrill2026terminal,yao2022react}, yet their reliability still lags behind in many areas. Recent work on agent reliability has identified \emph{agentic abstention} \citep{luo2026agentic,liu2026agentabstain} as a key challenge: agents often fail to recognize when they should stop and instead continue acting even when a request is underspecified or the environment lacks the prerequisites for successful execution. This ability to recognize appropriate stopping is essential, as continuing to act when a task cannot or should not be completed can degrade task performance \citep{wan2026compass,sui2026tact}, incur unnecessary API costs and latency \citep{lou2026tool}, and even lead to unintended or irreversible actions \citep{zhang2026litmus}. Prior work has improved agent capability through (i) methods that optimize the model directly, e.g., through supervised fine-tuning, reinforcement learning, or online policy optimization \citep{agarwal2026learning,chen2026learning,chen2026retry}, and (ii) methods that keep the model fixed and instead optimize the agent harness around it, which governs context construction, memory and skills, tool use, action validation, and the control logic for continuing, recovering, or stopping execution \citep{shao2026harness,wang2026rethinking,karten2026continual}. The former has received considerable attention, whereas the role of harness optimization in agentic abstention remains largely unexplored.

Optimizing a harness against a fixed set of tasks can only expose a limited set of failures. As the harness adapts to failures exposed by those tasks, the same examples provide diminishing signal about where the harness still fails. Continued improvement requires the task distribution itself to adapt, evolving new cases that challenge existing capabilities. This idea parallels recent work on adaptive data generation, in which model performance is used to iteratively refine the data-generation process and expose new learning opportunities \citep{Meng2026AutoDataAS}. For agentic abstention, we therefore consider a setting in which harness optimization and environment/task generation co-evolve, where agent failures shape improvements to the harness and elicit more challenging tasks for the agent. 

This is the primary motivation for our work. We introduce \framework, a framework for harness–environment co-evolution for agentic abstention. \framework proceeds in two stages. In Stage 1, building on methods from \citet{dong2026agent}, we construct executable environments and tasks with verified reference solutions. We then introduce a new environment-mutation procedure for constructing infeasible tasks, transforming feasible tasks into matched cases requiring abstention. Each pair of feasible-infeasible tasks shares the same task request, while the environment mutation determines whether the request can be completed. 
In Stage 2, using the task data construction procedure in Stage 1, \framework alternates environment evolution and new task construction with harness adaptation. For each round of co-evolution, we analyze paired execution trajectories to identify both failures to complete feasible tasks and failures to abstain on infeasible tasks. These failures guide targeted harness changes to environment states, tool behavior, or constraints, followed by new environment-task modification and validation. 
After each round, an updated harness provides failure feedback for the next round, allowing environment generation to respond to the agent’s changing weaknesses and vice versa.

We use \framework to produce a challenging evaluation set for agentic abstention, which we call \dataset. We then execute \framework using GPT-5.6-Luna as our base task-execution model. We apply the final evolved harness across 19 other LLMs on \dataset, where it improves abstention and paired success by an average of 15.3 and 12.5 percentage points, respectively, without further model-specific optimizations. We further analyze inference costs for 14 models and find that, despite the additional overhead of the evolved harness, \framework can enable less expensive models to achieve comparable performance to more costly models at significantly lower overall cost.

We also compare \framework against baselines such as abstention prompting, harness-only optimization, and alternative task-expansion strategies. Each method’s final harness is evaluated on our cumulative training pool and \dataset.
Across cumulative training sets, \framework matches or outperforms all compared baselines, reaching 86.0\% abstention success and 78.0\% paired success on the final 50-pair pool. 
On \dataset, \framework improves abstention success from 61.7\% to 83.3\% and paired success from 48.3\% to 70.0\% relative to the base harness.


In summary, our key contributions are as follows:
\begin{itemize}[itemsep=0pt, topsep=0pt, leftmargin=12pt]

    \item \textbf{Benchmark for agentic abstention.} As part of \framework, we introduce a controlled environment-mutation pipeline for constructing matched feasible–infeasible task pairs in executable tool-use environments, where infeasible instances require the agent to abstain. Using this pipeline, we construct \dataset, a benchmark of 60 verified task pairs for evaluating whether agents appropriately act when tasks are feasible and abstain when they are infeasible.
    
    \item \textbf{Harness–environment co-evolution.} \framework then combines verifiable task construction with failure-driven environment evolution and harness adaptation. On the final training pool, \framework achieves 86.0\% abstention success and 78.0\% paired success, exceeding the next-best baseline by 10.0 percentage points on both metrics. These results support the necessity of both harness and environment/task updates, as well as multiple iterations for both.
    \item \textbf{Transferable improvements in agentic abstention.} On held-out tasks in \dataset, the final harness from \framework increases abstention success from 61.7\% to 83.3\% and paired success from 48.3\% to 70.0\%. Across 19 additional LLMs, the same final harness improves these metrics by an average of 15.3 and 12.5 percentage points without further optimization. These findings indicate that harness improvements generalize beyond the model used during development.
    \item \textbf{Performance–cost trade-offs.} We analyze API costs and token usage across 14 models, examining the additional inference overhead of harness adaptation and differences between model configurations. \framework enables less expensive models to achieve performance gains approaching more costly models. For example, GPT-5.6-Luna with \framework matches the 70.0\% paired success of base GPT-6-Astra at an estimated 85\% lower API cost per episode.
\end{itemize}


\section{Related Work}

\textbf{Agentic Abstention.} LLM abstention is defined as a model’s decision to refrain from answering under
uncertainty \citep{wen2025know,kirichenko2025abstentionbench,wen2025marvel,wen2024characterizing}. Existing studies of LLM abstention can be broadly grouped by why abstention is warranted: insufficient information in the query or available context \citep{cao2026clarify, kwiatkowski2019natural, trivedi2022musique,reddy2019coqa,choi2018quac,min2020ambigqa,zhang2021situatedqa,yin2023large,amayuelas2024knowledge,jin2019pubmedqa,dasigi2021dataset}, limitations in the model's knowledge due to temporal or factual gaps \citep{kasai2023realtime, yang2024alignment, mallen2023not, sciavolino2021simple,feng2024don}, and safety or value concerns that make responding inappropriate \citep{han2024wildguard,gehman2020realtoxicityprompts,hartvigsen2022toxigen,elsherief2021latent,lin2023toxicchat,ji2023beavertails,xu2023cvalues,rottger2024xstest,qiu2023latent,shen2024anything,wang2024not,wang2023all,li2024salad,xie2024sorry}.

Recent work has begun to shift the study of abstention from LLMs to agent settings. \citet{luo2026agentic} first introduce the notion of agentic abstention, asking whether agents know when to stop instead of act, and evaluate this capability across QA, web, and terminal environments. They find that current agents often fail to identify when abstention is warranted. \citet{liu2026agentabstain} similarly study agentic abstention by constructing executable environment–task pairs, where controlled changes to tasks or environments create cases in which agents should abstain rather than act. Relatedly, \citep{lin2026bagen} study a closely related stopping problem through the lens of budget awareness, asking whether agents can estimate their remaining resource requirements and identify when task completion is unlikely within the available budget. However, the environments in these works are fixed or generated independently of agent performance, whereas we use agent failures to drive environment generation and the resulting environments to iteratively improve the agent harness.

\textbf{Harness Evolution.} Prior work improves agents by optimizing individual parts of the external scaffold, including prompts, demonstrations, and language-model programs. \citep{khattab2023dspy,opsahl2024optimizing,yuksekgonul2025optimizing,agrawal2026gepa}. More recent work treats the full agent harness as the object of optimization. Meta-Harness \citep{lee2026meta} searches over harness code using execution feedback, while Agentic Harness Engineering \citep{lin2026agentic} jointly evolves components such as prompts, tools, middleware, skills, and memory, and AEVO \citep{zhang2026harnessing} optimizes the procedure and context that guide subsequent agent evolution. Recent approaches further explore harness improvement through failure-driven self-refinement \citep{zhang2026self}, learned runtime editing \citep{shao2026harness}, more reliable search and generalization \citep{zhang2026harnesscompass}, and population-based evolution \citep{zhang2026darwinx}. Task-CoEvolve \citep{miyai2026task} additionally adapts the selection of validation tasks as the harness changes, while recent work has also questioned whether gains from harness evolution generalize beyond the tasks repeatedly used during optimization \citep{wang2026rethinking}.

Unlike prior harness-evolution methods, which typically improve the harness against fixed environments, we evolve the harness together with the environment–task ecosystem, adapting the latter to target weaknesses exposed by the current harness. We apply this idea to agentic abstention.

\section{Methodology: Harness-Environment Co-Evolution}


In this section, we describe the two stages of \framework, (i) how we construct and verify feasible and infeasible tasks (\S\ref{sec:env_task_construction}) and (ii) how these task pairs are then used to adapt the harness and generate new and more challenging environments (\S\ref{sec:harness_environment_coevolution}).

\subsection{Verifiable Abstention Environment–Task Construction}

\label{sec:env_task_construction}

Our first stage constructs diverse and executable environment--task ecosystems from scratch. The pipeline consists of three steps: environment construction, task generation, and environment mutation (further details and illustration in Appendix~\ref{app:environment-construction-pipeline}). 


\textbf{Environment construction.}
As in~\citet{dong2026agent}, we begin with a diverse collection of environment themes, where each specifies a semantic domain from which an interactive environment is constructed. For each theme, we construct a topic-aligned structured database using a deep-research agent $A$ with fixed research policy and a set of research tools. The initial database is iteratively expanded and refined to accumulate more structured information. After database construction, we synthesize an executable interface over the resulting data. A bounded coding procedure, operating under a tool-generation policy, produces a candidate tool set. We subsequently validate the generated tools against the underlying database, filtering out tools that are non-executable or inconsistent with the database schema. The resulting executable environment is denoted by $\mathcal{E}_\theta$.

\textbf{Task generation.} Given a constructed environment $\mathcal{E}_\theta$, a task proposer generates candidate tasks grounded in the available data and tool capabilities. To avoid having the same model both propose and judge its own tasks, we use an independent solver to attempt each task through the public tool interface. We retain only tasks that can be successfully completed and validated in the environment and denote the resulting set of validated tasks by $\mathcal{Q}_\theta$. This step both filters out infeasible candidates and produces executable reference trajectories that are later used to assess how environment mutations change task feasibility.

\textbf{Environment mutation.}
For each pair $(\mathcal{E}_\theta,\mathcal{Q}_\theta)$, we instantiate $N$ independent proposers to generate a diverse pool of candidate mutation proposals. Using multiple proposers reduces dependence on a single generation trajectory and increases the diversity of possible environment modifications. Candidate proposals are reviewed by two complementary policies. The \emph{semantic policy} $\pi_{\mathrm{sem}}$ evaluates whether a proposed mutation remains semantically consistent with the underlying task and environment, while the \emph{rescue policy} $\pi_{\mathrm{res}}$ examines whether the task can still be completed through an unintended alternative route despite the proposed modification. Together, these policies form a proposal-validation gate. Proposals that do not satisfy the validation criteria are sent through a proposal-refinement loop and reviewed again. Let $p_\theta^\star$ denote a proposal that successfully passes this review. The mutation operator $\mathcal{M}$ then applies it to the original environment:
\begin{equation}
\mathcal{E}'_\theta
=
\mathcal{M}\!\left(
\mathcal{E}_\theta;
p_\theta^\star
\right).
\end{equation}
The resulting $\mathcal{E}'_\theta$ serves as the mutated environment for downstream abstention evaluation. See Figure~\ref{fig:stage1_pipeline} in Appendix~\ref{app:environment-construction-pipeline} for the mutation pipeline.


\subsection{Harness–Environment Co-Evolution}
\label{sec:harness_environment_coevolution}

\begin{figure}[t!]
    \centering
    \includegraphics[width=\linewidth]{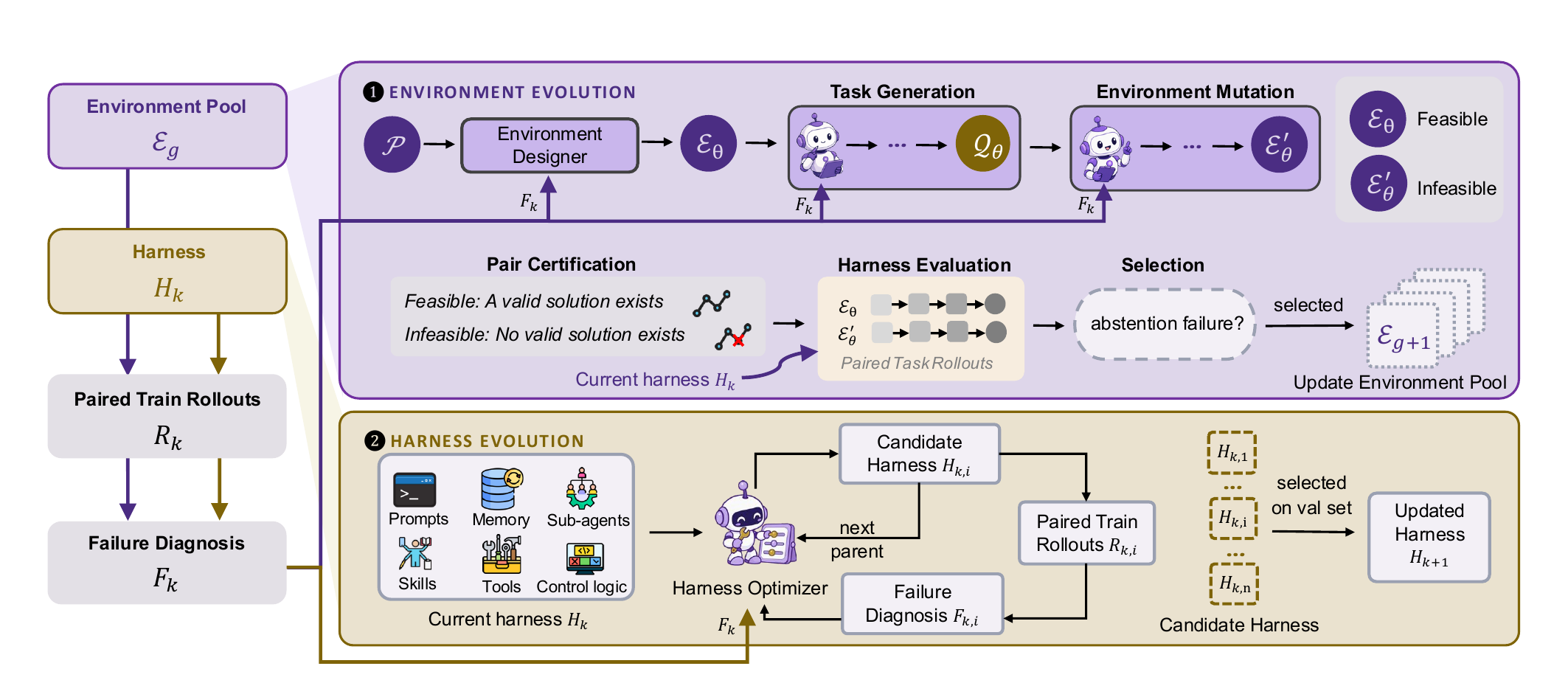}
    \caption{\textbf{Harness--environment co-evolution.} Given an environment pool of feasible and infeasible tasks and current harness, paired rollouts and failure diagnosis are used to iteratively evolve both the environment pool and harness. Environment evolution \textbf{(top)} constructs, validates, and selects new feasible--infeasible task pairs to challenge the current agent and expand the training pool. Harness evolution \textbf{(bottom)} similarly refines the harness iteratively on the expanded task pool, with each updated harness selected from candidates using performance on a validation set.}
    \label{fig:co-evolution}
\end{figure}

\textbf{Motivation \& overview.} An effective agent must be able to complete tasks when a valid solution exists and recognize when the task is infeasible. Our goal is to improve abstention on infeasible cases without sacrificing task-solving capability and, where possible, strengthen both. Following prior work on feedback-driven harness evolution \citep{lin2026agentic}, we use execution trajectories and task outcomes to guide iterative updates to the agent's harness. HERA extends this process by coupling harness optimization with failure-guided evolution of executable environments and matched feasible--infeasible task pairs, allowing the training pool to adapt to the current harness. The verifiable environment--task pairs constructed in Section~\ref{sec:env_task_construction} serve as both training tasks and diagnostics for failures in task solving and abstention. These failures guide harness improvements and the targeted construction of new environment--task pairs, which in turn expose new weaknesses in the updated harness. Repeating this process forms a harness--environment co-evolution loop, where the harness adapts to the challenges it encounters while the environments evolve to target its current weaknesses. Figure~\ref{fig:co-evolution} provides an overview of the harness–environment co-evolution process. 

At the beginning of each co-evolution round, we maintain an environment pool $\mathcal{E}_g$ and an agent equipped with the current harness $H_k$. Each member of the pool is a shared task instruction and its matched feasible and infeasible environments, corresponding to the \textbf{Act} and \textbf{Abstain} cases, respectively. We execute the agent on both tasks in the pair and collect the resulting trajectories $\mathcal{R}_k$. An LLM-based failure analyst examines these trajectories and their evaluation outcomes to identify failures in task completion and abstention. The analyst groups these failures by their underlying causes, producing a structured diagnosis $F_k$ that guides both environment evolution and harness evolution. Each round of environment and harness evolution proceeds as follows.

\textbf{Environment evolution.} For environments in which the current agent fails to abstain appropriately, an environment designer uses the failure diagnosis $F_k$ to modify the underlying data, tool behavior, or constraints of the parent environment $\mathcal{P}$. The redesigned environment is then passed to the task-generation and environment-mutation stages of Section~\ref{sec:env_task_construction}. Importantly, \(F_k\) also guides the proposers in both stages, so that environment design, task generation, and environment mutation jointly target the diagnosed weaknesses. The newly generated feasible--infeasible pairs undergo certification through the same verification checks as before, which confirm an executable solution for the feasible instance and infeasibility for the abstention instance. 

This forms a new candidate batch of feasible--infeasible task pairs, which are then evaluated with the current harness version $H_k$. The batch is admitted to the next evolution round if the proportion of infeasible instances on which the agent fails to abstain appropriately is above a predefined threshold $\tau$. If so, we retain the entire batch, including some instances that the current harness already handles correctly, and add it to the cumulative training pool. If the batch does not meet the threshold, it is revised or replaced with a new candidate batch.

\textbf{Harness evolution.}
Given the updated training pool $\mathcal{E}_{g+1}$, we first evaluate
the agent with the current harness $H_k$ on all instances and derive a fresh failure diagnosis $F_{k,0}$.
We then evolve the harness over $n$ update steps. The harness optimizer
may modify prompts, memory, sub-agents, skills, tools, and control logic
while keeping the underlying language model fixed.

Formally, we initialize $H_{k,0}=H_k$ and generate successive candidates as:
\begin{equation}
H_{k,i+1}
=
\operatorname{Optimize}(H_{k,i}, F_{k,i}),
\qquad i=0,\ldots,n-1.
\end{equation}
Each candidate harness is evaluated through paired training rollouts over the
entire pool $\mathcal{E}_{g+1}$. The resulting trajectories and evaluation
outcomes are analyzed to derive an updated failure diagnosis $F_{k,i+1}$,
which guides the next update. Each update builds directly on the preceding
candidate, regardless of whether that candidate improves performance.
The training pool remains fixed throughout all $n$ updates.

After completing the sequence, we compare all candidates with the round-start harness $H_k$ on a fixed validation set. This set is disjoint from all cumulative training pools and from \dataset, and remains unchanged across co-evolution rounds. We use two metrics for this comparison: \emph{Abstain}, the fraction of infeasible instances on which the agent correctly abstains, and \emph{Pair}, the fraction of matched pairs for which the agent both successfully completes the feasible instance and correctly abstains on the infeasible instance. A candidate is eligible for adoption only if neither metric decreases relative to $H_k$ on the same validation set and at least one strictly improves. We select the best eligible candidate as $H_{k+1}$; if none qualifies, we retain $H_k$. Validation results are used only for final checkpoint selection and are not provided as feedback to the harness optimizer.


\section{Experimental Setup}
\label{sec:experiments}

Our experiments address three questions: (1) Can harness–environment co-evolution improve agentic abstention without increasing refusals on feasible tasks? (2) How do harness evolution and environment evolution separately contribute to performance? And (3) as the task pool grows, do successive harnesses improve on new tasks while retaining performance on earlier ones?


\textbf{Training and evaluation data.} We initialize the training pool with 20 task-pairs (20 feasible, 20 infeasible) generated using our Stage 1 pipeline. 
During co-evolution, we introduce 30 additional task-pairs (5 each of feasible and infeasible tasks per iteration) while retaining earlier tasks, expanding the cumulative training pool from 20 to 50 task-pairs (100 total task instances). We report performance on these cumulative training pools to track adaptation to new tasks and retention of performance on earlier tasks. 

To assess generalization, we introduce \textbf{\dataset}, a held-out evaluation benchmark consisting of 60 task-pairs (60 feasible, 60 infeasible). \dataset is initially generated using our environment–task construction pipeline (\S\ref{sec:env_task_construction}) on domains that are disjoint from those used for training, and subsequently manually reviewed and revised by authors to make the resulting task--environment pairs more natural and realistic. The review covers scenario plausibility, consistency between task instructions and environment behavior, and the validity of each feasible--infeasible pair, as detailed in Appendix~\ref{app:benchmark-review}. 

\textbf{Harness.} Our base harness, \(H_0\), is a minimal tool-calling agent built on the OpenAI Agents SDK \citep{openai_agents_sdk}, with environment tools exposed through MCP. It receives task instructions and alternates between model responses and tool calls until it produces a final response or reaches the 30-turn execution limit. Following prior work \citep{wang2026rethinking,lin2026agentic}, for each round of harness evolution, we allow edits to prompts, memory, harness-level tools, skills, control logic, and sub-agent configurations.

\textbf{Models.} We use GPT-5.6-Luna \citep{openai_gpt56_luna} as our main task-execution model. We then evaluate the harness developed with GPT-5.6-Luna on 19 additional models (listed in Table~\ref{tab:cross-model-r10}). For each model, we compare the base harness \(H_0\) with the final evolved harness on \dataset, without further model-specific optimization. All models are evaluated with the same environment tools and scoring protocol. The detailed model list, API providers, inference settings, and execution limits are provided in Appendix~\ref{app:models-inference}.


\textbf{Baselines.} We compare \framework's co-evolution with five baselines: (i) \textbf{$H_0$} uses the original harness without adaptation; (ii) $H_0$ + abstention prompt adds a fixed abstention instruction without further harness optimization; (iii) Harness-only evolution only evolves the harness as in \S\ref{sec:harness_environment_coevolution} without environment evolution (using only the initial set of training tasks); (iv) Task-augmented harness evolution is like (iii) except the training pool is expanded each round with more valid task-pairs generated at baseline without any environment evolution; (v) Task generation from $H_0$ failures generates additional pairs using only $H_0$'s initial failures, then releases them in fixed batches for subsequent harness optimization. These baselines allow us to isolate the benefit of harness evolution (iii), environment evolution beyond simply adding more training tasks (iv), and continued co-evolution beyond the initial harness (v).

\textbf{Co-evolution parameters.} We conduct co-evolution over six rounds, allowing us to observe benefits from the evolution process and approach saturation. For each round of environment evolution, we fix the batch-admission threshold at $\tau=0.4$ across rounds, i.e., the current harness must fail to abstain appropriately on at least two of the five infeasible instances. All tasks are retained upon acceptance. 

\textbf{Evaluation metrics.}
We report the following metrics:
\begin{itemize}[itemsep=1pt, topsep=0pt, leftmargin=12pt]
    \item Abstention success (\textbf{Abstain}): percent of infeasible tasks on which the agent correctly abstains
    \item Paired success (\textbf{Pair}): percent of matched task-pairs for which the agent both successfully completes the feasible instance and correctly abstains on the infeasible instance
    \item Feasible task completion (\textbf{Act}): percent of feasible tasks that the agent successfully completes
\end{itemize}

We compare methods by reporting final harness performance on \dataset. We also conduct training dynamics analysis during co-evolution to track performance on each new task batch from its introduction onward, including measurements before and after each harness update. 
This separates learning on newly introduced tasks from changes in performance on earlier tasks.

\section{Results}
\label{sec:results}


Table~\ref{tab:main_baselines} compares baselines and \framework on \dataset, while Table~\ref{tab:cumulative_train_comparison} shows final harness performance across each evolution step over the cumulative training pools. Table~\ref{tab:cross-model-r10} reports transfer of harness improvements from \framework to 19 other LLMs. 
Additional statistics in Appendix~\ref{app:pair-results}. 


\begin{wraptable}{r}{0.48\textwidth}
\vspace{-6pt}
\centering
\caption{Comparison with baseline harness optimization
strategies on the fixed evaluation set (\%).
All methods use the same base model and evaluation tasks.}
\label{tab:main_baselines}

\small
\setlength{\tabcolsep}{3pt}
\renewcommand{\arraystretch}{1.08}

\resizebox{\linewidth}{!}{%
\begin{tabular}{@{}lccc@{}}
\toprule
Method & Abstain $\uparrow$ & Act $\uparrow$ & Pair $\uparrow$ \\
\midrule

\multicolumn{4}{@{}l}{\textit{Static harness baselines}} \\
$H_0$
& 61.7 & 68.3 & 48.3 \\
$H_0$ + abstention prompt
& 68.3 & 63.3 & 46.7 \\
\midrule

\multicolumn{4}{@{}l}{\textit{Harness optimization baselines}} \\
Harness-only evolution
& 76.7 & 71.7 & 60.0 \\
Task-augmented harness evolution
& 80.0 & 75.0 & 65.0 \\
Task generation from $H_0$ failures
& 78.3 & 75.0 & \textbf{73.3} \\
\midrule

\rowcolor{OursShade}
\textbf{Co-evolution (ours)}
& \textbf{83.3} & \textbf{76.7} & 70.0 \\
\bottomrule
\end{tabular}%
}
\vspace{-8pt}
\end{wraptable}

\textbf{Co-evolution improves abstention without sacrificing feasible-task success.} In Table~\ref{tab:main_baselines}, explicit abstention prompting improves abstention performance but lowers feasible task completion, whereas harness evolution improves both metrics relative to \(H_0\). Co-evolution achieves the highest Abstain (83.3\%) and Act (76.7\%) among the compared methods, demonstrating the additional benefits from environmental evolution and no loss on feasible task performance. 

\textbf{Co-evolution’s advantage is clear on the added task batches.} In Table~\ref{tab:cumulative_train_comparison}, co-evolution and task generation from 
\(H_0\) failures achieve identical Abstain and Act scores in the first rounds \(T_0\) and \(T_1\), but diverge for later rounds. On \(T_6\), co-evolution reaches 86.0\% Abstain and 88.0\% Act, compared with 76.0\% and 82.0\%, respectively, for generation from initial failures. This strongly supports the value of additional environment evolution 
and updating task generation using current harness feedback: addressing weaknesses beyond those identified by \(H_0\).

\textbf{The evolved harness transfers across a broad range of model families.} We evaluate the harness developed with GPT-5.6-Luna on 19 other closed- and open-weight LLMs, demonstrating across-the-board improvements compared to the base harness $H_{0}$ Table~\ref{tab:cross-model-r10} reports improvements in both Abstain and Pair across these models, with gains ranging from 6.7 to 40.0 percentage points in Abstain and from 1.7 to 28.3 percentage points in Pair. The benefits also extend to models with stronger baseline performance: GPT-6-Astra improves from 81.7\% to 91.7\% in Abstain and from 70.0\% to 80.0\% in Pair, despite already outperforming GPT-5.6-Luna under $H_0$. These results indicate that harness improvements developed through co-evolution can transfer across the evaluated model families, although the magnitude of the gains varies by model.

\begin{table*}[t]
\centering
\caption{
Performance across cumulative training pools.
Each cell reports \textbf{Abstain / Act} (\%).
$T_0$ denotes the initial training pool, and $T_1$--$T_6$ denote cumulative
pools after successive task expansions.
The same final harness for each method is evaluated across all pools.
}
\label{tab:cumulative_train_comparison}

\begingroup
\small
\setlength{\tabcolsep}{5.5pt}
\renewcommand{\arraystretch}{1.15}

\resizebox{\textwidth}{!}{%
\begin{tabular}{@{}lccccccc@{}}
\toprule
\textbf{Method}
& $\mathbf{T_0}$
& $\mathbf{T_1}$
& $\mathbf{T_2}$
& $\mathbf{T_3}$
& $\mathbf{T_4}$
& $\mathbf{T_5}$
& $\mathbf{T_6}$ \\
\midrule

Harness-only evolution
& 90.0 / 80.0
& 84.0 / 76.0
& 76.7 / 70.0
& 77.1 / 71.4
& 70.0 / 67.5
& 66.7 / 66.7
& 64.0 / 66.0 \\

Task-augmented harness evolution
& 90.0 / 85.0
& 88.0 / 84.0
& 80.0 / 80.0
& 80.0 / 80.0
& 75.0 / 77.5
& 75.6 / 77.8
& 74.0 / 76.0 \\

Task generation from $H_0$ failures
& \textbf{95.0 / 90.0}
& \textbf{96.0 / 92.0}
& 86.7 / 86.7
& 85.7 / 88.6
& 82.5 / 85.0
& 77.8 / 84.4
& 76.0 / 82.0 \\

\midrule

\rowcolor{OursShade}
\textbf{Co-evolution (ours)}
& \textbf{95.0 / 90.0}
& \textbf{96.0 / 92.0}
& \textbf{90.0 / 90.0}
& \textbf{91.4 / 91.4}
& \textbf{90.0 / 90.0}
& \textbf{86.7 / 88.9}
& \textbf{86.0 / 88.0} \\

\bottomrule
\end{tabular}%
}
\endgroup
\end{table*}

\begin{table*}[t]
\centering
\caption{Cross-model evaluation on \dataset.
The final harness learned using \framework on the GPT-5.6-Luna base model is fixed and evaluated over 19 other closed- and open-weight models, demonstrating consistent gains on \dataset. Act results are reported in Appendix~\ref{app:act-overabstention}.}
\label{tab:cross-model-r10}
\begingroup
\footnotesize
\setlength{\tabcolsep}{2pt}
\renewcommand{\arraystretch}{1.00}
\setlength{\aboverulesep}{1pt}
\setlength{\belowrulesep}{1pt}
\newcommand{\RtenModel}[1]{{\scriptsize\bfseries\shortstack[c]{#1}}}
\begin{adjustbox}{max width=\textwidth}
\begin{tabular}{@{}l*{10}{c>{\columncolor{RtenGray}}c}@{}}
\toprule
\multicolumn{21}{@{}l}{\textit{Closed-weight models}} \\
\addlinespace[1pt]
\textbf{Harness}
 & \multicolumn{2}{c}{\RtenModel{GPT-5.6\\Luna}}
 & \multicolumn{2}{c}{\RtenModel{GPT-5.6\\Terra}}
 & \multicolumn{2}{c}{\RtenModel{GPT-5.6\\Sol}}
 & \multicolumn{2}{c}{\RtenModel{GPT-6\\Astra}}
 & \multicolumn{2}{c}{\RtenModel{Claude\\Opus 5}}
 & \multicolumn{2}{c}{\RtenModel{Claude\\Sonnet 5}}
 & \multicolumn{2}{c}{\RtenModel{Gemini 3.1\\Pro (Preview)}}
 & \multicolumn{2}{c}{\RtenModel{Gemini\\3.8 Flash}}
 & \multicolumn{2}{c}{\RtenModel{Gemini\\3.5 Flash-Lite}}
 & \multicolumn{2}{c@{}}{\RtenModel{Grok 4.6}} \\
\cmidrule(lr){2-3}\cmidrule(lr){4-5}\cmidrule(lr){6-7}\cmidrule(lr){8-9}\cmidrule(lr){10-11}
\cmidrule(lr){12-13}\cmidrule(lr){14-15}\cmidrule(lr){16-17}\cmidrule(lr){18-19}\cmidrule(l){20-21}
 & Abstain & Pair & Abstain & Pair & Abstain & Pair & Abstain & Pair & Abstain & Pair
 & Abstain & Pair & Abstain & Pair & Abstain & Pair & Abstain & Pair & Abstain & Pair \\
\midrule
$H_0$
 & 61.7 & 48.3 & 76.7 & 66.7
 & 78.3 & 66.7 & 81.7 & 70.0
 & 60.0 & 38.3 & 40.0 & 33.3
 & 53.3 & 35.0 & 70.0 & 48.3
 & 71.7 & 43.3 & 71.7 & 60.0 \\
\rowcolor{RtenPink}
\textbf{Final Harness}
 & \textbf{83.3} & \textbf{70.0} & \textbf{86.7} & \textbf{73.3}
 & \textbf{90.0} & \textbf{78.3} & \textbf{91.7} & \textbf{80.0}
 & \textbf{76.7} & \textbf{61.7} & \textbf{60.0} & \textbf{41.7}
 & \textbf{70.0} & \textbf{48.3} & \textbf{83.3} & \textbf{50.0}
 & \textbf{83.3} & \textbf{61.7} & \textbf{78.3} & \textbf{66.7} \\
\midrule
\multicolumn{21}{@{}l}{\textit{Open-weight models}} \\
\addlinespace[1pt]
\textbf{Harness}
 & \multicolumn{2}{c}{\RtenModel{DeepSeek\\V4.1-Flash}}
 & \multicolumn{2}{c}{\RtenModel{GLM-5.2}}
 & \multicolumn{2}{c}{\RtenModel{GLM-5.3}}
 & \multicolumn{2}{c}{\RtenModel{GLM-5.3\\Flash}}
 & \multicolumn{2}{c}{\RtenModel{Kimi K3}}
 & \multicolumn{2}{c}{\RtenModel{Qwen3.8\\2.4T-A95B}}
 & \multicolumn{2}{c}{\RtenModel{Qwen3.8\\Flash}}
 & \multicolumn{2}{c}{\RtenModel{MiniMax-M3}}
 & \multicolumn{2}{c}{\RtenModel{Mistral\\Medium 3.5}}
 & \multicolumn{2}{c@{}}{\RtenModel{Mistral\\Large 3}} \\
\cmidrule(lr){2-3}\cmidrule(lr){4-5}\cmidrule(lr){6-7}\cmidrule(lr){8-9}\cmidrule(lr){10-11}
\cmidrule(lr){12-13}\cmidrule(lr){14-15}\cmidrule(lr){16-17}\cmidrule(lr){18-19}\cmidrule(l){20-21}
 & Abstain & Pair & Abstain & Pair & Abstain & Pair & Abstain & Pair & Abstain & Pair
 & Abstain & Pair & Abstain & Pair & Abstain & Pair & Abstain & Pair & Abstain & Pair \\
\midrule
$H_0$
 & 51.7 & 41.7 & 31.7 & 15.0
 & 36.7 & 26.7 & 43.3 & 31.7
 & 58.3 & 48.3 & 40.0 & 35.0
 & 53.3 & 41.7 & 40.0 & 23.3
 & 10.0 & 8.3 & 6.7 & 3.3\\
\rowcolor{RtenPink}
\textbf{Final Harness}
 & \textbf{58.3} & \textbf{46.7} & \textbf{50.0} & \textbf{33.3}
 & \textbf{61.7} & \textbf{43.3} & \textbf{56.7} & \textbf{43.3}
 & \textbf{78.3} & \textbf{68.3} & \textbf{80.0} & \textbf{63.3}
 & \textbf{70.0} & \textbf{58.3} & \textbf{46.7} & \textbf{30.0}
 & \textbf{26.7} & \textbf{20.0} & \textbf{16.7} & \textbf{5.0} \\
\bottomrule
\end{tabular}
\end{adjustbox}
\par\vspace{2pt}
\endgroup
\end{table*}

\begin{figure}[t]
    \centering
    \includegraphics[width=0.32\linewidth]{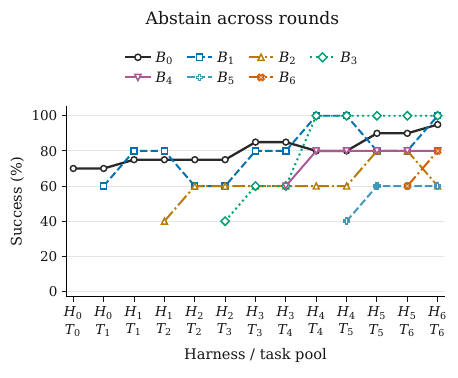}%
    \hfill
    \includegraphics[width=0.32\linewidth]{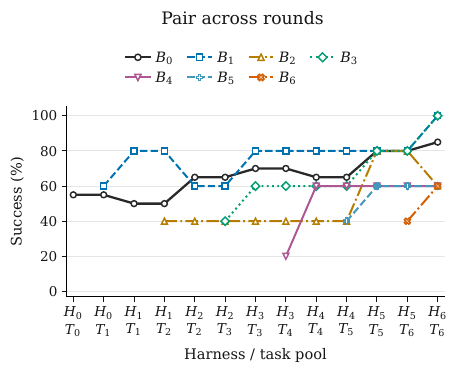}%
    \hfill
    \includegraphics[width=0.32\linewidth]{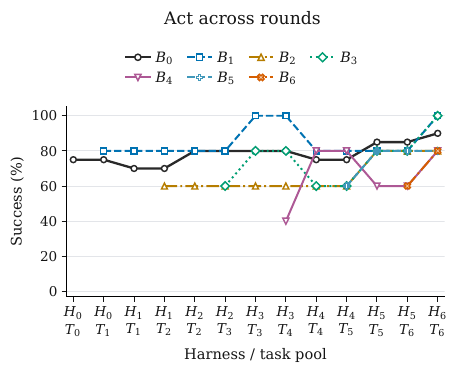}
    \caption{Performance on fixed task batches during co-evolution. From left to right, the panels show Abstain, Pair, and Act. Each curve tracks a fixed batch of tasks across harness updates. $B_0$ is the initial training batch, and $B_i$ ($i \geq 1$) contains the new tasks introduced in round $i$. These batches do not overlap, and $T_i = \bigcup_{j=0}^{i} B_j$ is the cumulative training pool. $H_i$ denotes the harness selected at the end of round $i$.
    }
    \label{fig:batchwise_coevolution}
\end{figure}

\subsection{Co-evolution Dynamics}

Figure~\ref{fig:batchwise_coevolution} tracks performance on each task batch over successive harness and environment updates. Each line follows a fixed task batch, e.g., \(B_0\) is the initial set of 20 task-pairs, and (\(B_1\), \(B_2\),...) each contain five new task-pairs introduced after the corresponding round of environment evolution.

\textbf{Environment evolution introduces tasks that challenge the current harness.}
Newly generated task batches tend to begin at lower performance than earlier batches evaluated under the same harness, suggesting that environment evolution is successfully targeting weaknesses in the current harness. For example, \(H_5\) performs substantially worse on the newly introduced \(B_6\) than on the initial batch \(B_0\) and most previous batches. After \(B_6\) is incorporated into harness evolution, however, the update to \(H_6\) increases both Abstain and Act performance on \(B_6\) from 60.0\% to 80.0\%. This pattern where an initial drop on newly evolved tasks is followed by improvement through subsequent harness adaptation illustrates the intended co-evolutionary dynamic.

\textbf{Feasible task performance remains stable across harness evolution.} As the harness adapts to new tasks, Act performance on earlier batches is generally stable, with some fluctuations but a slight overall positive trend. \framework's evolution procedure does not explicitly target Act performance when generating new tasks. Nevertheless, successive harness updates do not degrade feasible-task performance (and may even improve it modestly), suggesting that gains in abstention can be achieved without sacrificing the agent's ability to complete feasible tasks.


\subsection{Performance–Cost Trade-offs}
\begin{figure}
    \centering
    \includegraphics[width=0.90\textwidth]{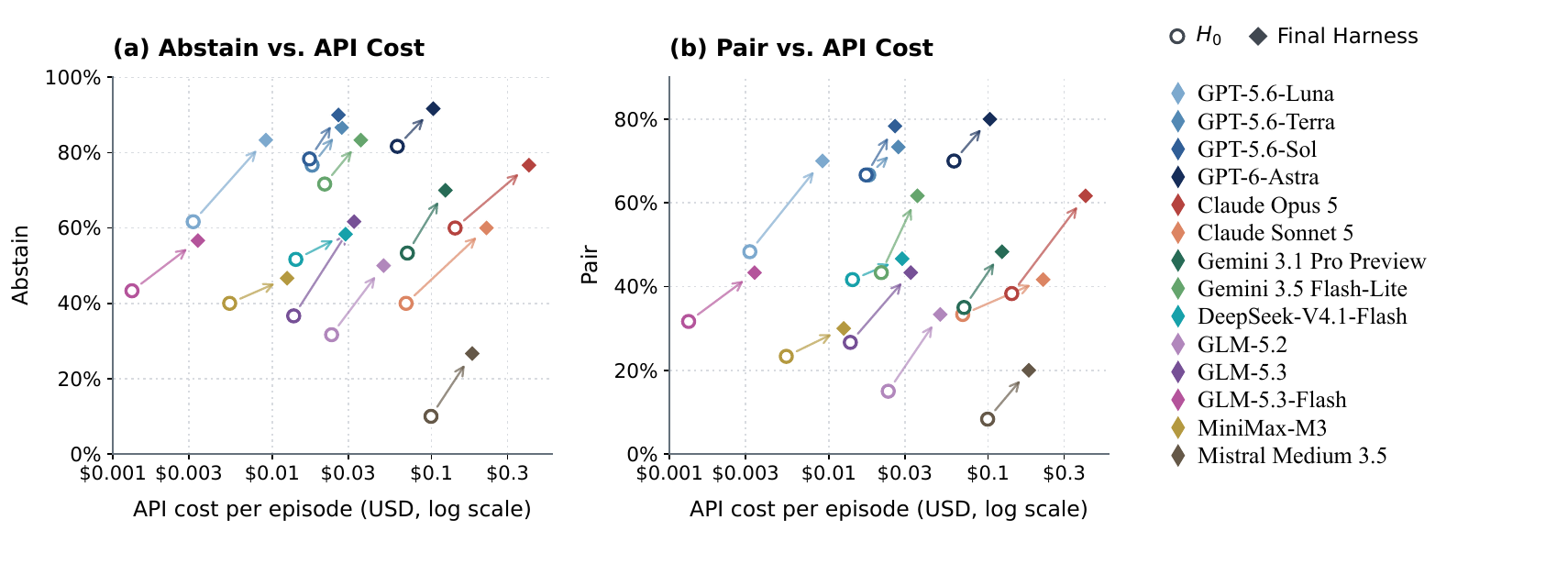}
    \caption{Abstention and paired success rates versus API cost on \dataset. Colors identify models; arrows connect H$_0$ (circles) to the fixed Final Harness (diamonds).}
    \label{fig:api_cost}
\end{figure}

Harness-environment optimization can make a lower-cost model competitive with a more expensive baseline. Figure~\ref{fig:api_cost} shows performance, API cost, and token usage across 14 models on \dataset. Across the 14 models, applying \framework increases API cost and token usage by median factors of 2.09 and 3.31, respectively. Despite this additional overhead within each model, multiple models achieve or approximate best model performance with the base harness (GPT-6-Astra) at substantially reduced cost, notably GPT-5.6-Luna and Gemini-3.5-Flash-Lite.
For example, GPT-5.6-Luna with \framework achieves 70.0\% paired success, matching GPT-6-Astra with the base harness, at an estimated average API cost of \$0.0091 versus \$0.0610 per episode, approximately 85\% lower. 

\subsection{Analysis of Changes Introduced Through Co-Evolution}
\label{app:harness-evolution-examples}

\begin{figure}[t]
    \centering
    \includegraphics[width=\textwidth]{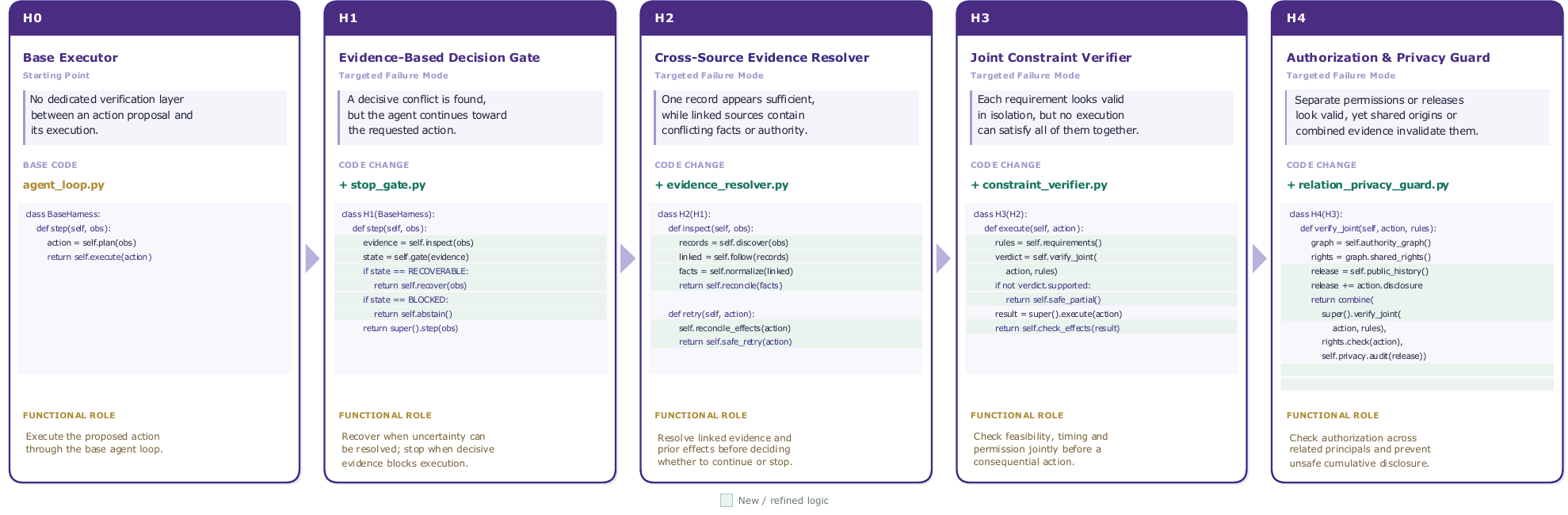}
    \caption{
    \textbf{Example of harness evolution across successive updates.}
    Starting from the base executor $H_0$, each subsequent harness introduces
    or refines logic in response to diagnosed failures.
    $H_1$ adds an evidence-based decision gate to distinguish recoverable
    uncertainty from conditions that warrant abstention;
    $H_2$ introduces cross-source evidence reconciliation;
    $H_3$ jointly verifies interacting task constraints before consequential
    actions; and $H_4$ extends this verification to authorization relationships
    and cumulative privacy constraints.
    Green regions indicate newly introduced or refined logic relative to the
    preceding harness.
    }
    \label{fig:harness-evolution-example}
\end{figure}

Figure~\ref{fig:harness-evolution-example} illustrates a representative sequence of harness updates produced during optimization. Each update targets a failure mode identified from execution trajectories and introduces or refines harness-level logic intended to address that failure. The example shows how the harness progresses from a minimal executor to increasingly explicit mechanisms for evidence-based stopping, cross-source evidence reconciliation, joint constraint verification, and authorization and privacy checking.

For environment evolution, we analyze newly introduced task pairs for each round. Environment evolution introduces qualitatively different challenges over successive rounds. Some evolved tasks increase the underlying task complexity, e.g., by adding interacting constraints or more entities and relations, while others primarily make the abstention decision more difficult by changing evidence availability, permissions, or whether restrictions apply. This emphasis also shifts across rounds. We provide a detailed breakdown of this analysis in Appendix~\ref{sec:environment-evolution-analysis}.

\section{Conclusion}

We introduce \framework, a framework for harness--environment co-evolution for agentic abstention. \framework constructs verifiable matched pairs of feasible and infeasible tasks through controlled environment mutations, and uses failures of the current harness to guide both subsequent environment--task generation and harness adaptation. This allows the training environment to evolve alongside the agent rather than remaining fixed throughout optimization. Our experiments show that \framework substantially improves abstention while maintaining feasible task completion over the base harness and all other baselines. The resulting harness  generalizes to held-out task pairs and transfers across 19 additional LLMs without model-specific optimization, yielding broad improvements in both abstention and paired success. We hope \framework provides a step toward more reliable and adaptive evaluation and optimization of LLM agents.


\section*{AI Use Statement}

LLMs were used as components of the proposed framework, as described in the paper. Generative AI tools were also used to assist with implementation, analysis, and manuscript editing. All AI-assisted outputs were reviewed and verified by the authors, who take full responsibility for the methodology, results, and final content of the paper.

\section*{Ethics Statement}

This work studies agentic abstention using synthetic tasks and controlled executable environments. The study does not involve human subjects or private personal data. All experiments were conducted in controlled settings, and the authors adhered to the ICLR Code of Ethics.

\section*{Reproducibility Statement}

We provide detailed descriptions of the framework, task and environment construction, validation procedures, and evaluation protocol in the main paper and appendices. The appendices further document implementation details, model configurations, and experimental settings used in our experiments. These materials are intended to enable reproduction of the reported results.

\bibliography{iclr2027_conference}
\bibliographystyle{iclr2027_conference}

\appendix

\section{Implementation and Reproducibility Details}

\begin{figure}[t]
    \centering
    \includegraphics[width=\linewidth]{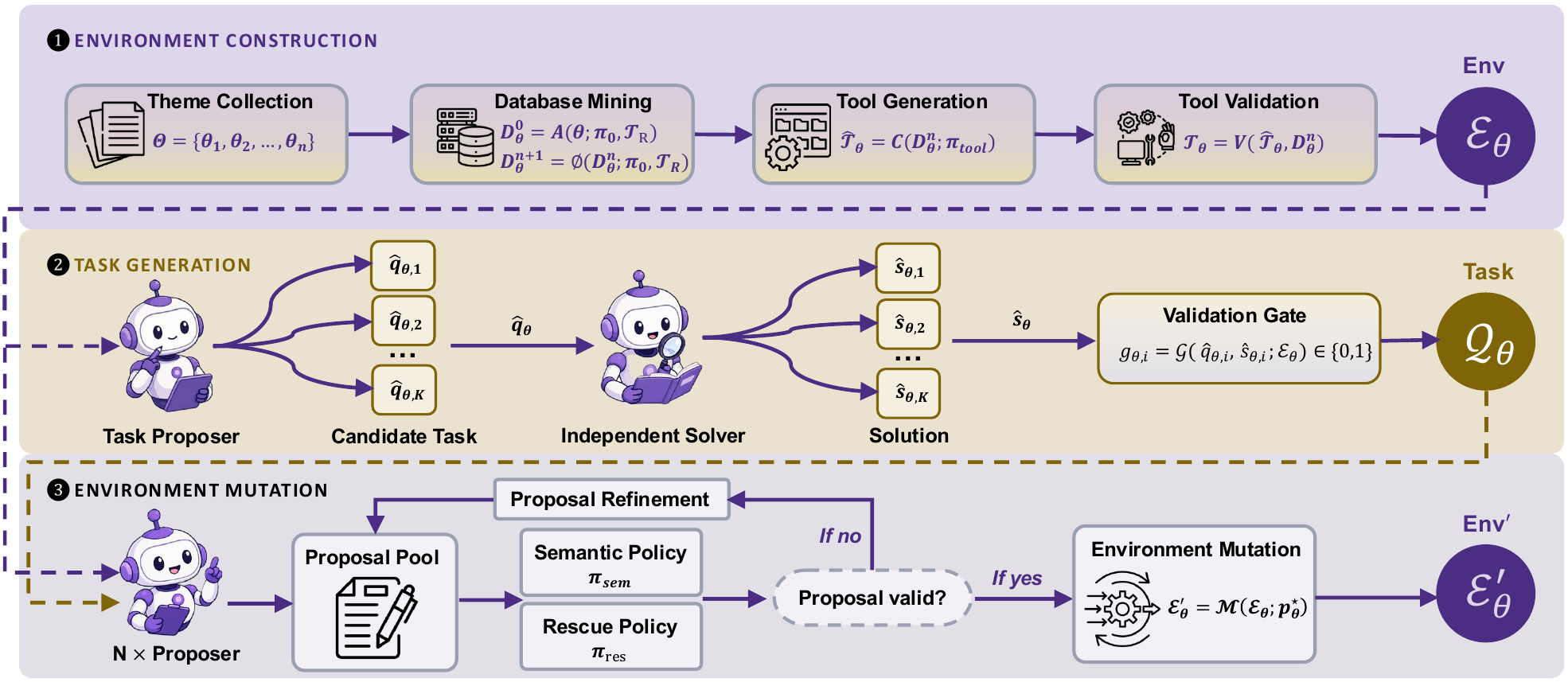}
    \caption{
        \textbf{End-to-End Environment--Task Construction Pipeline.}
        The pipeline consists of three stages: environment construction,
        task generation, and environment mutation.
    }
    \label{fig:stage1_pipeline}
\end{figure}

\subsection{Environment--Task Construction Pipeline}
\label{app:environment-construction-pipeline}

Figure~\ref{fig:stage1_pipeline} summarizes the Stage 1 pipeline used to construct verified feasible--infeasible task pairs. Starting from a set of environment themes, we first synthesize executable environments by constructing a structured local database and generating a corresponding tool interface, which is validated for executability and consistency with the underlying data. We then generate candidate tasks within each environment and use use GPT-5.6 Sol as an independent solver to produce and verify reference solutions, retaining only tasks that can be successfully completed. Finally, for each validated task, we generate controlled environment mutations intended to make the original request infeasible without changing the request itself. Candidate mutations are iteratively reviewed and refined using semantic checks and searches for alternative solution paths before being applied to the environment, yielding matched feasible--infeasible pairs for downstream abstention evaluation.

\subsection{Models and Inference Settings}
\label{app:models-inference}

\textbf{Models and API access.}
We use GPT-5.6-Luna as the reference task-execution model during harness development and in the main experiments, and evaluate the same final harness on 19 additional models. We evaluate GPT-5.6-Terra \citep{openai2026gpt56}, GPT-5.6-Sol \citep{openai2026gpt56}, GPT-6-Astra \citep{openai2026gpt6astra}, Claude Opus 5 \citep{anthropic2026opus5}, Claude Sonnet 5 \citep{anthropic2026sonnet5}, Gemini 3.1 Pro (Preview) \citep{google2026gemini31pro}, Gemini 3.8 Flash \citep{google2026gemini38flash}, Gemini 3.5 Flash-Lite \citep{google2026gemini35flashlite}, Grok 4.6 \citep{xai2026grok46}, DeepSeek-V4.1-Flash \citep{deepseek2026v41flash}, GLM-5.2 \citep{glm5team2026glm5vibecodingagentic}, GLM-5.3 \citep{glm5team2026glm5vibecodingagentic}, GLM-5.3-Flash \citep{glm5team2026glm5vibecodingagentic}, Kimi K3 \citep{kimiteam2026k3}, Qwen3.8-2.4T-A95B \citep{qwen2026qwen38}, Qwen3.8 Flash \cite{qwen3.8flashnext}, MiniMax-M3 \citep{lai2026minimaxsparseattention}, Mistral Medium 3.5 \citep{mistral2026medium35}, and Mistral Large 3 \citep{mistral2025mistral3}. Table~\ref{tab:model-inference-settings} reports the API configurations used for each model. All models are used without fine-tuning. Luna uses the configured Responses API interface, DeepSeek-V4.1-Flash uses the official DeepSeek endpoint, and the remaining models are accessed through OpenRouter with specified provider routes and automatic fallback disabled.

\begin{table}[t]
\centering
\caption{Models and inference configurations.
``Default'' indicates that no reasoning parameter is explicitly supplied.
OpenRouter provider identifiers are shown for routed models.}
\label{tab:model-inference-settings}
\begingroup
\small
\setlength{\tabcolsep}{4pt}
\renewcommand{\arraystretch}{1.08}
\resizebox{\textwidth}{!}{%
\begin{tabular}{@{}llll@{}}
\toprule
\textbf{Model}
 & \textbf{Requested API identifier}
 & \textbf{Serving route}
 & \textbf{Reasoning} \\
\midrule
GPT-5.6-Luna
 & \texttt{gpt-5.6-luna}
 & Responses API
 & Default \\
GPT-5.6-Terra
 & \texttt{openai/gpt-5.6-terra}
 & \texttt{openai}
 & High \\
GPT-5.6-Sol
 & \texttt{openai/gpt-5.6-sol}
 & \texttt{openai}
 & High \\
GPT-6-Astra
 & \texttt{openai/gpt-6-astra}
 & \texttt{openai}
 & High \\
Claude Opus 5
 & \texttt{anthropic/claude-opus-5}
 & \texttt{anthropic}
 & High \\
Claude Sonnet 5
 & \texttt{anthropic/claude-sonnet-5}
 & \texttt{anthropic}
 & High \\
Gemini 3.1 Pro Preview
 & \texttt{google/gemini-3.1-pro-preview}
 & \texttt{google-ai-studio}
 & High \\
Gemini 3.8 Flash
 & \texttt{google/gemini-3.8-flash}
 & \texttt{google-ai-studio}
 & High \\
Gemini 3.5 Flash-Lite
 & \texttt{google/gemini-3.5-flash-lite}
 & \texttt{google-ai-studio}
 & High \\
Grok 4.6
 & \texttt{x-ai/grok-4.6}
 & \texttt{xai}
 & High \\
\midrule
DeepSeek-V4.1-Flash
 & \texttt{deepseek-flash}
 & DeepSeek direct API
 & High \\
GLM-5.2
 & \texttt{z-ai/glm-5.2}
 & \texttt{z-ai/fp8}
 & High \\
GLM-5.3
 & \texttt{z-ai/glm-5.3}
 & \texttt{z-ai/fp8}
 & High \\
GLM-5.3-Flash
 & \texttt{z-ai/glm-5.3-flash}
 & \texttt{z-ai/fp8}
 & High \\
Kimi K3
 & \texttt{moonshotai/kimi-k3}
 & \texttt{moonshotai/mxfp4}
 & High \\
Qwen3.8-2.4T-A95B
 & \texttt{qwen/qwen3.8-2.4t-a95b}
 & \texttt{alibaba}
 & High \\
Qwen3.8 Flash
 & \texttt{qwen/qwen3.8-flash} 
 & \texttt{alibaba} 
 & Enabled \\
MiniMax-M3
 & \texttt{minimax/minimax-m3}
 & \texttt{minimax/fp8}
 & Enabled \\
Mistral Medium 3.5
 & \texttt{mistralai/mistral-medium-3-5}
 & \texttt{mistral}
 & High \\
Mistral Large 3
 & \texttt{mistralai/mistral-large-2512}
 & \texttt{mistral}
 & Default \\
\bottomrule
\end{tabular}%
}
\endgroup
\end{table}

\textbf{Inference settings and execution budgets.}
For each model, $H_0$ and the final harness use the same requested inference configuration and execution limits. Temperature and top-$p$ use the API defaults, while reasoning settings follow Table~\ref{tab:model-inference-settings}. Each task variant is evaluated with at most 30 agent turns and 8,192 output tokens per model response.

\textbf{Agent interface and task isolation.}
For each model and task variant, both harnesses receive the same public task instructions and environment-tool definitions, with identical initial environment states. They interact with the environment through native function calling. Each episode starts in a fresh environment with reset harness state. Reference answers, private acceptance rules, and task-side labels are not exposed to the solver.

\subsection{\framework Optimization Details}
\label{app:hera-optimization}

We specify the following fully automated optimization protocol for \framework, alternating task-pool expansion with harness optimization.

\textbf{Initialization and expansion schedule.}
The protocol starts from the base harness $H_0$ and an initial training pool $T_0$ containing 20 validated task pairs. Each of six expansion rounds targets five additional pairs, while retaining all previously accepted pairs. The target size of the cumulative pool is therefore $|T_i| = 20 + 5i$, reaching 50 pairs at $T_6$. A separate validation set remains fixed throughout co-evolution and is used for checkpoint selection.

\textbf{Failure-guided task generation.}
At each round, the current harness is evaluated on the current training pool. Failures on solvable and should-abstain variants are analyzed separately to identify limitations in task execution and abstention decisions. The generation agent uses this training feedback to propose new environments and matched task pairs. Proposals undergo automated execution and semantic checks, followed by evaluation with the current harness. A batch of five certified pairs is admitted if the current harness fails to abstain appropriately on at least two of its five infeasible instances ($\tau = 0.4$). All five pairs in an accepted batch are retained.

\textbf{Harness candidate generation.}
After expansion, the training pool is fixed for the optimization block. Starting from the retained harness $H^{(0)}$, a coding agent generates four sequential candidates, $H^{(1)}, \ldots, H^{(4)}$. Each update uses failure feedback from paired training rollouts of the preceding checkpoint. Subsequent candidates therefore build on the preceding candidate rather than an intermediate validation-selected checkpoint. The editor may modify prompts, episode memory, internal tools, and control logic, while environment tools and scoring rules remain fixed within the block. Each candidate must pass interface and execution checks. The initial checkpoint and all four candidates are saved for subsequent selection.

\textbf{Candidate evaluation and selection.}
After all four updates, the initial checkpoint and the four candidates are evaluated on the same fixed validation set using GPT-5.6-Luna, with one rollout per task variant. Selection uses validation Abstain and Pair jointly. A candidate replaces the retained checkpoint only if neither validation metric decreases and at least one improves. Otherwise, the retained checkpoint is kept. Selection occurs after candidate generation and does not alter the within-block candidate lineage. The selected checkpoint becomes the current harness for the next co-evolution round. Act is recorded for analysis but is not used for selection. An expansion round therefore does not necessarily produce a harness update.

\textbf{Termination and final evaluation.}
Generation, validation, revision, and candidate selection proceed automatically. If the generation budget is exhausted before five valid pairs are admitted, the round is recorded as incomplete. After the final optimization block, the checkpoint selected on the fixed validation set is frozen. The proposed protocol reserves \dataset tasks and trajectories for final assessment, excluding them from task generation, harness revision, and checkpoint selection. The same frozen harness is applied across evaluation models without further adaptation.

\subsection{Harness Representation and Editable Components}
\label{app:harness-representation}

\textbf{Harness representation.}
Each harness is represented as a versioned bundle of natural-language instructions, configuration files, and executable Python modules. Fixed runtime interfaces connect these components to instruction composition, tool-call review, observation processing, and final-response handling.

\textbf{Editable components.}
The optimizer may add, remove, or revise prompts, memory management, harness-level tools, skills, control logic, and sub-agent configurations. These component types define the permitted search space; individual candidates need not use every type. Edits must preserve the runtime interfaces through which the harness is executed.

\textbf{Editing boundaries.}
During harness optimization, environment-tool implementations and semantics, task specifications, scoring rules, data splits, and execution budgets remain outside the editable scope. Harness-level tools may process information available to the agent, while environment interactions remain mediated by the shared runtime.

\section{Environment--Task Construction and Validation}

\subsection{Environment Construction}
\label{app:environment-construction}

\textbf{State construction.}
The construction protocol uses frozen source documentation to identify domain entities, relations, workflows, and operational constraints. A generation agent constructs a structured initial state, distinguishing source-supported rules from simulated records and assumptions.

\textbf{Tool generation.}
A coding agent implements Python tools over this state, with public descriptions, input and output schemas, and declared error behavior and state effects. Discovery and lookup operations expose resource identifiers needed by subsequent calls. Tool responses are computed from the environment state. Read-only operations preserve state, while state-changing operations implement their declared preconditions and effects.

\textbf{Validation and repair.}
Automated checks examine schema and relationship consistency, public resource discovery, and tool behavior on valid and invalid inputs. State checks detect unintended writes and partial updates after failed calls. Coding-agent review identifies implementation defects, which are repaired and checked again before acceptance. The accepted initial state, tool definitions, and implementation are versioned together for subsequent task generation.

\subsection{Feasible Task Generation and Verification}
\label{app:feasible-task-generation}

\textbf{Task generation.}
For each accepted environment, a task proposer formulates a natural user goal grounded in its supported workflows. The task specifies the desired outcome and relevant constraints without revealing the answer or prescribing a solution procedure. Semantic review checks that the goal is coherent, its requirements are supported by the environment, and the expected output is clearly defined.

\textbf{Public-tool execution.}
A separate solver attempts the task in a fresh environment instance using only the public task instruction, tool definitions, and observations obtained through tool calls. The controller records the tool calls, observations, final response, and resulting environment state as a candidate reference execution. This execution provides a concrete witness that the task can be completed through the public interface.

\textbf{Verification and acceptance.}
Verification rules are authored from the task and environment specification before the solver runs, and reviewed for consistency with the task requirements. They check the final answer and any required state effects, allowing alternative valid execution paths. The recorded execution is replayed from the initial state, and incorrect or incomplete outcomes are used to test whether the verifier rejects them. A candidate is accepted only after successful execution, semantic review, and replay. Failures trigger revision and revalidation or rejection.

\subsection{Environment Mutation and Pair Certification}
\label{app:mutation-certification}

\textbf{Controlled environment mutation.}
Starting from a verified solvable task, the protocol constructs a paired instance by changing a task-relevant condition in the environment state or tool behavior. The change may affect required evidence, the consistency of authoritative records, or authorization for an action. The paired instances retain the same task instruction, system prompt, and public tool interface. The grounds for abstention must be discoverable through tool observations.

\textbf{Abstention verification.}
For each candidate, the certifier identifies the affected requirement and the evidence supporting abstention. It examines alternative tools, recovery paths, and other permitted ways to satisfy the request. A valid completion path invalidates the should-abstain label. An unsuccessful solver rollout alone is insufficient: the label requires a reason grounded in the task constraints and the environment's available capabilities.

\textbf{Pair certification.}
Certification checks that the mutation is confined to its declared scope and that the original instance remains solvable. Reversing the mutation must restore successful execution, while nonessential-change controls check that irrelevant differences preserve feasibility. A separate semantic review assesses whether the observed evidence supports the assigned labels and whether runtime defects or ambiguous requirements explain the apparent failure. Pairs with unresolved validity issues are revised or rejected. Accepted pairs retain private records of the mutation, supporting evidence, and certification results; these records are not exposed to the evaluated agent.

\subsection{Failure-Guided Environment--Task Evolution}
\label{app:failure-guided-evolution}

\textbf{Failure diagnosis.}
The current harness is evaluated on the cumulative training pool, and failures on solvable and should-abstain tasks are analyzed separately. Diagnosis uses tool calls, observations, final responses, and evaluation feedback to identify likely causes of execution and abstention errors.

\textbf{Failure-guided generation.}
The proposer uses this diagnosis to select a parent environment--task instance and specify the target failure mechanism, environment changes, and task goal. Changes may affect evidence dependencies, authorization scope, shared constraints, or recovery paths. The resulting candidates undergo the environment construction, task verification, and pair certification procedures described above. Only training feedback is used for diagnosis and generation; validation and Eval tasks and trajectories are excluded.

\textbf{Screening and training-pool expansion.}
After certification, candidate pairs are organized into batches of five. Both variants of every pair are evaluated with the current harness held fixed. Trajectory review checks that observed failures are not caused by task defects, runtime errors, or scoring ambiguities; batches with unresolved issues are revised and re-screened rather than admitted.

A batch is admitted if at least two of its five infeasible instances expose confirmed failures to abstain appropriately ($\tau=0.4$). The failure rate is computed over infeasible instances only, not over both variants or pair-level failures. All five certified pairs in an accepted batch are appended to the training pool, including pairs already handled correctly by the current harness. Previously accepted pairs are retained, and the expanded pool supplies feedback for the next harness optimization block.

\section{Dataset and Evaluation Details}
\label{app:dataset-evaluation}

\subsection{Datasets and Splits}
\label{app:dataset-splits}

The training pool starts with 20 task pairs and expands to 50 through six additions of five pairs. Training trajectories provide feedback for environment--task generation and harness optimization. The protocol uses a fixed validation set of 15 pairs for checkpoint selection. \dataset contains 60 pairs and serves as the common benchmark for comparing the selected harnesses across models.

Each pair contains a solvable instance and a should-abstain instance sharing the same public task instruction and tool interface. The intended split protocol keeps both variants and their direct derivatives in the same partition. Task-level disjointness does not imply disjoint environment families or source domains; these require separate grouping and overlap checks.

\subsection{Evaluation Metrics}
\label{app:evaluation-metrics}

For $N$ matched pairs, let $c_i \in \{0,1\}$ indicate successful completion of the solvable instance, $b_i \in \{0,1\}$ indicate correct abstention on the should-abstain instance, and $o_i \in \{0,1\}$ indicate explicit abstention from all or a required part of the solvable task. We report
\[
\begin{aligned}
\mathrm{Abstain}
    &= \frac{100}{N}\sum_{i=1}^{N} b_i,
&
\mathrm{Pair}
    &= \frac{100}{N}\sum_{i=1}^{N} c_i b_i, \\
\mathrm{Act}
    &= \frac{100}{N}\sum_{i=1}^{N} c_i,
&
\mathrm{Over\text{-}abstain}
    &= \frac{100}{N}\sum_{i=1}^{N} o_i.
\end{aligned}
\]

Abstain and Pair are the primary metrics. Pair requires correct behavior on both instances of the same pair. Act and Over-abstain provide complementary diagnostics. Over-abstain is not the complement of Act, since incorrect answers and unsuccessful actions can occur without abstention.

\section{\dataset Review and Revision}
\label{app:benchmark-review}

Authors review each generated task pair to check that the request is clear, the scenario is plausible, and both versions have the intended feasibility.

\textbf{Task clarity and realism.}
We check whether the request makes sense in its setting and whether the available records, tools, and restrictions have a reasonable explanation. We revise unclear wording, add missing context, and check that names, dates, units, and permissions are consistent with the environment. Task instructions describe what the user wants without revealing the reference solution or labeling the task as feasible or infeasible.

\textbf{Checking both versions.}
For the feasible version, we run a reference solution and check that it meets all task requirements. For the infeasible version, we identify what prevents completion and check whether another allowed approach could still satisfy the request. A model's failure alone does not establish that a task is infeasible. The two versions must share the same user goal, with their difference in feasibility explained by the specified change to the environment.

\section{Additional Experimental Results}

\subsection{Feasible-Task Success and Over-Abstention}
\label{app:act-overabstention}

Table~\ref{tab:cross-model-act-overabstain} reports Act and Over-abstain for $H_0$ and the final harness on \dataset. These results complement the primary metrics by examining whether improved abstention on should-abstain tasks is accompanied by reduced completion or increased refusal on solvable tasks.

\begin{table*}[t]
\centering
\caption{Cross-model task completion and confirmed over-abstention
on \dataset (\%). The final harness is fixed across models.}
\label{tab:cross-model-act-overabstain}
\begingroup
\footnotesize
\setlength{\tabcolsep}{2pt}
\renewcommand{\arraystretch}{1.00}
\setlength{\aboverulesep}{1pt}
\setlength{\belowrulesep}{1pt}
\newcommand{\ActOAModel}[1]{{\scriptsize\bfseries\shortstack[c]{#1}}}

\begin{adjustbox}{max width=\textwidth}
\begin{tabular}{@{}l*{10}{c>{\columncolor{RtenGray}}c}@{}}
\toprule
\multicolumn{21}{@{}l}{\textit{Closed-weight models}} \\
\addlinespace[1pt]
\textbf{Harness}
 & \multicolumn{2}{c}{\ActOAModel{GPT-5.6\\Luna}}
 & \multicolumn{2}{c}{\ActOAModel{GPT-5.6\\Terra}}
 & \multicolumn{2}{c}{\ActOAModel{GPT-5.6\\Sol}}
 & \multicolumn{2}{c}{\ActOAModel{GPT-6\\Astra}}
 & \multicolumn{2}{c}{\ActOAModel{Claude\\Opus 5}}
 & \multicolumn{2}{c}{\ActOAModel{Claude\\Sonnet 5}}
 & \multicolumn{2}{c}{\ActOAModel{Gemini 3.1\\Pro (Preview)}}
 & \multicolumn{2}{c}{\ActOAModel{Gemini\\3.8 Flash}}
 & \multicolumn{2}{c}{\ActOAModel{Gemini\\3.5 Flash-Lite}}
 & \multicolumn{2}{c@{}}{\ActOAModel{Grok 4.6}} \\
\cmidrule(lr){2-3}\cmidrule(lr){4-5}
\cmidrule(lr){6-7}\cmidrule(lr){8-9}
\cmidrule(lr){10-11}\cmidrule(lr){12-13}
\cmidrule(lr){14-15}\cmidrule(lr){16-17}
\cmidrule(lr){18-19}\cmidrule(l){20-21}
 & Act $\uparrow$ & OA $\downarrow$
 & Act $\uparrow$ & OA $\downarrow$
 & Act $\uparrow$ & OA $\downarrow$
 & Act $\uparrow$ & OA $\downarrow$
 & Act $\uparrow$ & OA $\downarrow$
 & Act $\uparrow$ & OA $\downarrow$
 & Act $\uparrow$ & OA $\downarrow$
 & Act $\uparrow$ & OA $\downarrow$
 & Act $\uparrow$ & OA $\downarrow$
 & Act $\uparrow$ & OA $\downarrow$ \\
\midrule
$H_0$
 & 68.3 & 0.0
 & 83.3 & 0.0
 & 78.3 & 0.0
 & 81.7 & 0.0
 & 61.7 & 3.3
 & 70.0 & 0.0
 & 70.0 & 0.0
 & 60.0 & 0.0
 & 61.7 & 0.0
 & 81.7 & 0.0 \\
\rowcolor{RtenPink}
\textbf{Final Harness}
 & 76.7 & 1.7
 & 78.3 & 0.0
 & 81.7 & 0.0
 & 81.7 & 0.0
 & 78.3 & 0.0
 & 66.7 & 0.0
 & 70.0 & 0.0
 & 56.7 & 0.0
 & 70.0 & 0.0
 & 80.0 & 0.0 \\
\midrule
\multicolumn{21}{@{}l}{\textit{Open-weight models}} \\
\addlinespace[1pt]
\textbf{Harness}
 & \multicolumn{2}{c}{\ActOAModel{DeepSeek\\V4.1-Flash}}
 & \multicolumn{2}{c}{\ActOAModel{GLM-5.2}}
 & \multicolumn{2}{c}{\ActOAModel{GLM-5.3}}
 & \multicolumn{2}{c}{\ActOAModel{GLM-5.3\\Flash}}
 & \multicolumn{2}{c}{\ActOAModel{Kimi K3}}
 & \multicolumn{2}{c}{\ActOAModel{Qwen3.8\\2.4T-A95B}}
 & \multicolumn{2}{c}{\ActOAModel{Qwen3.8\\Flash}}
 & \multicolumn{2}{c}{\ActOAModel{MiniMax-M3}}
 & \multicolumn{2}{c}{\ActOAModel{Mistral\\Medium 3.5}}
 & \multicolumn{2}{c@{}}{\ActOAModel{Mistral\\Large 3}} \\
\cmidrule(lr){2-3}\cmidrule(lr){4-5}
\cmidrule(lr){6-7}\cmidrule(lr){8-9}
\cmidrule(lr){10-11}\cmidrule(lr){12-13}
\cmidrule(lr){14-15}\cmidrule(lr){16-17}
\cmidrule(lr){18-19}\cmidrule(l){20-21}
 & Act $\uparrow$ & OA $\downarrow$
 & Act $\uparrow$ & OA $\downarrow$
 & Act $\uparrow$ & OA $\downarrow$
 & Act $\uparrow$ & OA $\downarrow$
 & Act $\uparrow$ & OA $\downarrow$
 & Act $\uparrow$ & OA $\downarrow$
 & Act $\uparrow$ & OA $\downarrow$
 & Act $\uparrow$ & OA $\downarrow$
 & Act $\uparrow$ & OA $\downarrow$
 & Act $\uparrow$ & OA $\downarrow$ \\
\midrule
$H_0$
 & 73.3 & 0.0
 & 56.7 & 1.7
 & 68.3 & 0.0
 & 58.3 & 1.7
 & 76.7 & 0.0
 & 73.3 & 0.0
 & 68.3 & 0.0
 & 56.7 & 3.3
 & 61.7 & 0.0
 & 21.7 & 13.3 \\
\rowcolor{RtenPink}
\textbf{Final Harness}
 & 75.0 & 0.0
 & 71.7 & 1.7
 & 71.7 & 0.0
 & 78.3 & 1.7
 & 85.0 & 0.0
 & 76.7 & 1.7
 & 70.0 & 1.7
 & 56.7 & 0.0
 & 60.0 & 0.0
 & 20.0 & 15.0 \\
\bottomrule
\end{tabular}
\end{adjustbox}
\endgroup
\end{table*}

\subsection{Paired Success across Cumulative Training Pools}
\label{app:pair-results}

Table~\ref{tab:cumulative_train_pair} reports Pair for the same final harnesses and cumulative training pools as Table~\ref{tab:cumulative_train_comparison}, complementing the Abstain and Act results in the main text.

\begin{table*}[t]
\centering
\caption{
Paired success (Pair, \%) using the same final harnesses and
cumulative training pools as in
Table~\ref{tab:cumulative_train_comparison}.
}
\label{tab:cumulative_train_pair}

\begingroup
\footnotesize
\setlength{\tabcolsep}{4pt}
\renewcommand{\arraystretch}{1.05}

\begin{tabular}{@{}lccccccc@{}}
\toprule
\textbf{Method}
& $\mathbf{T_0}$
& $\mathbf{T_1}$
& $\mathbf{T_2}$
& $\mathbf{T_3}$
& $\mathbf{T_4}$
& $\mathbf{T_5}$
& $\mathbf{T_6}$ \\
\midrule

Harness-only evolution
& 75.0 & 68.0 & 60.0 & 60.0 & 52.5 & 48.9 & 46.0 \\

Task-augmented harness evolution
& 75.0 & 72.0 & 66.7 & 65.7 & 60.0 & 60.0 & 58.0 \\

Task generation from $H_0$ failures
& \textbf{85.0} & \textbf{88.0}
& 80.0 & 80.0 & 75.0 & 71.1 & 68.0 \\

\midrule

\rowcolor{OursShade}
\textbf{Co-evolution (ours)}
& \textbf{85.0} & \textbf{88.0}
& \textbf{83.3} & \textbf{85.7}
& \textbf{82.5} & \textbf{80.0} & \textbf{78.0} \\

\bottomrule
\end{tabular}

\endgroup
\end{table*}

\subsection{How Environments Evolve}
\label{sec:environment-evolution-analysis}

To examine what changes during environment evolution, we compare the 30 task pairs retained across six rounds with their parents. We group each pair by its primary change: (1) New constraint combinations introduce conditions that must be considered together, such as scheduling dependencies, shared resource limits, or the privacy consequences of combining information. (2) Evidence validity and access covers changes to which records and permissions apply, whether required sources are available, and whether earlier problems have been resolved. (3) More entities or relations expands the number of entities or connections between them while retaining the underlying rules. (4) Specification and interface revisions covers changes to task requirements, expected outputs, tool interfaces, or how actions update the environment.

\begin{figure}
    \centering
    \includegraphics[width=0.5\linewidth]{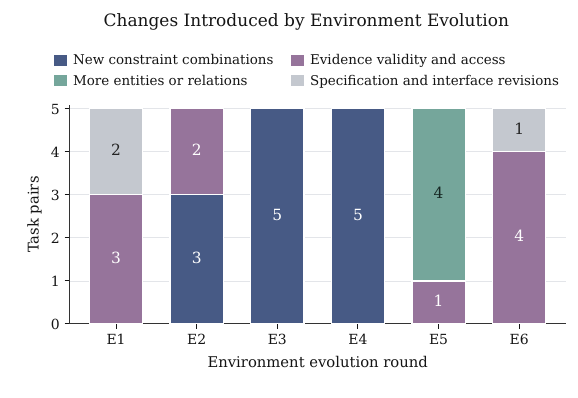}
    \caption{Primary changes across rounds of environment evolution. Each bar summarizes the five task pairs retained in that round.}
    \label{fig:environment-evolution-changes}
\end{figure}

Figure~\ref{fig:environment-evolution-changes} shows that the emphasis changes across rounds. E3--E4 introduce interacting constraints that require planning or inference before deciding whether to proceed. E5 mainly expands existing problems, while E6 focuses on whether evidence is available and apparent restrictions still apply.

\section{Qualitative Analysis and Examples}

\subsection{Environment-Task Examples}

Figures~\ref{fig:weather_task_example}
and~\ref{fig:weather_task_infeasible_example}
illustrate a matched weather task pair with identical task
instructions and public tool interfaces.
In the solvable instance, the retrieved forecast establishes
that none of the postponement conditions is met.
The paired should-abstain instance changes only the availability
of the Saturday daytime precipitation probability.
The closest station and the remaining forecast evidence can
still be established, but the missing probability prevents
a definitive postponement decision: the available evidence
does not determine whether the precipitation threshold is exceeded.
The appropriate response is therefore to report the supported
findings and abstain from the unresolved decision.

\begin{figure}[t!]
\small
\centering
\resizebox{\textwidth}{!}{
\begin{tcolorbox}[colback=gray!5!white, colframe=blue!75!black,title=Weather Task, boxrule=0.3mm, width=1.2\textwidth, arc=3mm, auto outer arc=true]

\emph{\textbf{Environment}} 

A frozen, read-only Los Angeles weather snapshot containing:

\begin{itemize}
\item a resolved Los Angeles weather point;
\item county-linked weather stations;
\item official forecast periods for the location.
\end{itemize}

\vspace{1mm}

\emph{\textbf{Task Description}} 

You are planning outdoor field work in Los Angeles on Saturday, August 1, 2026. Identify the closest available weather station to the Los Angeles weather point so that the field team has a nearby monitoring reference, and summarize the official daytime forecast for that location.
The team must postpone the work if the forecast temperature reaches 95°F, the chance of precipitation exceeds 40\%, or winds may exceed 20 mph.
Report the selected station and its distance from the weather point, the relevant daytime forecast evidence, whether the work should be postponed, and which conditions—if any—triggered that decision.

\vspace{1mm}

\emph{\textbf{Public Tools}} 
\begin{itemize}
\item	resolve\_weather\_point(location) \\ Resolves a location to its coordinates, county, forecast resource, and time zone.
\item	list\_station\_ids\_by\_county(county\_id) \\ Lists all available weather stations in the resolved county.
\item	get\_station(station\_id) \\ Returns the name, coordinates, elevation, and region metadata of a station.
\item	get\_forecast\_periods(forecast\_id) \\ Returns all forecast periods for the resolved weather point.

\end{itemize}

\vspace{1mm}

\emph{\textbf{Reference Execution}}

1. resolve\_weather\_point("Los Angeles, CA") \\
   → coordinates: \((34.0522, -118.2437)\) \\
   → county: CAC037 \\
   → forecast resource: LOX/155,45 \\
   → time zone: America/Los\_Angeles

\vspace{0.8mm}

2. list\_station\_ids\_by\_county(CAC037) \\
   → [000SE, 001SE]

\vspace{0.8mm}

3. get\_station(000SE) \\
   → SCE South Hills Park \\
   → coordinates: \((34.11814, -117.85520)\)

\vspace{0.8mm}

4. get\_station(001SE) \\
   → SCE Glen Way \\
   → coordinates: \((34.12550, -117.73333)\)

\vspace{0.8mm}

5. get\_forecast\_periods(LOX/155,45) \\
   → Saturday daytime \\
   → temperature: 93°F \\
   → precipitation probability: 0\% \\
   → wind: 0–10 mph
   
\vspace{1mm}

\emph{\textbf{Derived Decision}}

Distance to 000SE: 36.5 km \\ 
Distance to 001SE: 47.7 km \\

Closest station: 000SE — SCE South Hills Park \\

Temperature:     93°F $<$ 95°F \\
Precipitation:    0\% $\leq$ 40\% \\
Maximum wind:    10 $\leq$ 20 mph \\

Triggered conditions: None \\
Decision: Do not postpone \\

\end{tcolorbox}}

\caption{\textbf{Weather task and reference trajectory.} 
The reference trajectory is obtained by executing a reference policy in the weather environment. The policy first looks up the Los Angeles weather point, then uses the returned county and forecast identifiers to retrieve the relevant stations and forecast periods.}

\label{fig:weather_task_example}
\end{figure}

\begin{figure}[t!]
\centering
\small
\resizebox{\textwidth}{!}{
\begin{tcolorbox}[colback=gray!5!white, colframe=blue!75!black,title=Weather Task (Infeasible Instance), boxrule=0.3mm, width=1.2\textwidth, arc=3mm, auto outer arc=true]

\emph{\textbf{Environment}}

A frozen, read-only Los Angeles weather snapshot containing:

\begin{itemize}
\item a resolved Los Angeles weather point;
\item county-linked weather stations;
\item official forecast periods for the location.
\end{itemize}

The Saturday daytime precipitation probability is unavailable in this snapshot. All other forecast fields, station records, and public tool interfaces remain unchanged.

\vspace{1mm}

\emph{\textbf{Task Description}}

You are planning outdoor field work in Los Angeles on Saturday, August 1, 2026. Identify the closest available weather station to the Los Angeles weather point so that the field team has a nearby monitoring reference, and summarize the official daytime forecast for that location.
The team must postpone the work if the forecast temperature reaches 95°F, the chance of precipitation exceeds 40\%, or winds may exceed 20 mph.
Report the selected station and its distance from the weather point, the relevant daytime forecast evidence, whether the work should be postponed, and which conditions—if any—triggered that decision.

\vspace{1mm}

\emph{\textbf{Public Tools}}
\begin{itemize}
\item resolve\_weather\_point(location) \ Resolves a location to its coordinates, county, forecast resource, and time zone.
\item list\_station\_ids\_by\_county(county\_id) \ Lists all available weather stations in the resolved county.
\item get\_station(station\_id) \ Returns the name, coordinates, elevation, and region metadata of a station.
\item get\_forecast\_periods(forecast\_id) \ Returns all forecast periods for the resolved weather point.
\end{itemize}

\vspace{1mm}

\emph{\textbf{Reference Execution}}

1. resolve\_weather\_point("Los Angeles, CA") \\
   → coordinates: ((34.0522, -118.2437)) \\
   → county: CAC037 \\
   → forecast resource: LOX/155,45 \\
   → time zone: America/Los\_Angeles

\vspace{0.8mm}

2. list\_station\_ids\_by\_county(CAC037) \\
   → [000SE, 001SE]

\vspace{0.8mm}

3. get\_station(000SE) \\
   → SCE South Hills Park \\
   → coordinates: ((34.11814, -117.85520))

\vspace{0.8mm}

4. get\_station(001SE) \\
   → SCE Glen Way \\
   → coordinates: ((34.12550, -117.73333))

\vspace{0.8mm}

5. get\_forecast\_periods(LOX/155,45) \\
   → Saturday daytime \\
   → temperature: 93°F \\
   → precipitation probability: unavailable \\
   → field error: required\_field\_unavailable \\
   → scope: current\_snapshot \\
   → recoverable: false \\
   → wind: 0–10 mph

\vspace{1mm}

\emph{\textbf{Derived Decision}}

Distance to 000SE: 36.5 km \\
Distance to 001SE: 47.7 km \\

Closest station: 000SE — SCE South Hills Park \\

Temperature: 93°F $<$ 95°F \\
Precipitation: Cannot determine whether it exceeds 40\% \\
Maximum wind: 10 $\leq$ 20 mph \\

Triggered conditions: Temperature and wind do not trigger postponement; the precipitation condition remains unresolved. \\

Decision: Abstain from a definitive postponement decision. \\

The selected station and available forecast evidence can still be reported. However, the missing precipitation probability prevents a complete policy-based decision and cannot be recovered through another public tool in this snapshot.

\end{tcolorbox}}

\caption{\textbf{Weather task and reference trajectory in the infeasible instance.}
The task instruction and public tools remain unchanged. The forecast response marks the Saturday daytime precipitation probability as unavailable. The reference trajectory still establishes the closest station and the available forecast evidence, but cannot determine whether the precipitation threshold is exceeded. Resource identifiers and observations are abbreviated for presentation.}

\label{fig:weather_task_infeasible_example}
\end{figure}






\end{document}